\documentclass[11pt,reqno,final]{article}
\usepackage{amsthm}
\usepackage{amsmath,amsfonts,amssymb,amsthm,version}
\usepackage{mathrsfs,fancybox,pifont}
\usepackage{graphicx}
\usepackage{url,hyperref}
\usepackage[notcite,notref]{showkeys}
\usepackage{color}
\usepackage{multirow}
\usepackage{epstopdf}
\usepackage{cases}
\usepackage{mathtools}
\usepackage{algorithm,algorithmic}
\usepackage{authblk}
\usepackage{fancyhdr}
\usepackage{lipsum}
\usepackage{cite}
\usepackage{float}
\usepackage[top=2cm]{geometry}
\usepackage{booktabs}
\usepackage{subcaption}  
\usepackage[table]{xcolor}

\allowdisplaybreaks

\numberwithin{equation}{section}
\numberwithin{figure}{section}
\numberwithin{table}{section}

\theoremstyle{plain}

\theoremstyle{definition}

\theoremstyle{remark}

\title{Deep Learning for Automated Identification of Vietnamese Timber Species: A Tool for Ecological Monitoring and Conservation}
\author[1]{Tianyu Song } 
\author[2,3]{Van-Doan Duong } 
\author[2]{Thi-Phuong Le} 
\author[1]{Ton Viet Ta}

\affil[1]{Mathematical Modeling Laboratory, Graduate School of Bioresource and Bioenvironmental Sciences, Kyushu University}
\affil[2]{International Cooperation Center, Thai Nguyen University}
\affil[3]{Centrer of Crop Research for Adaptation to Climate Change, Thai Nguyen University of Agriculture and Forestry }

\date{}

\begin{document}
\maketitle

\begin{abstract}
Accurate identification of wood species plays a critical role in ecological monitoring, biodiversity conservation, and sustainable forest management. Traditional classification approaches relying on macroscopic and microscopic inspection are labor-intensive and require expert knowledge. In this study, we explore the application of deep learning to automate the classification of ten wood species commonly found in Vietnam. A custom image dataset was constructed from field-collected wood samples, and five state-of-the-art convolutional neural network  architectures—ResNet50, EfficientNet, MobileViT, MobileNetV3, and ShuffleNetV2—were evaluated. Among these, ShuffleNetV2 achieved the best balance between classification performance and computational efficiency, with an average accuracy of 99.29\% and F1-score of 99.35\% over 20 independent runs. These results demonstrate the potential of lightweight deep learning models for real-time, high-accuracy species identification in resource-constrained environments. Our work contributes to the growing field of ecological informatics by providing scalable, image-based solutions for automated wood classification and forest biodiversity assessment.


\end{abstract}

\section{Introduction} \label{introduction}
Wood, a versatile and renewable natural resource, has been an integral part of human civilization for millennia. Its wide-ranging applications span from architectural construction, furniture making, and musical instrument crafting, to energy generation and decorative art \cite{Zobel1989}. The proper classification and identification of wood species are essential across many of these fields, not only to ensure the correct use of timber for specific structural or aesthetic purposes, but also for legal, ecological, and conservation-related considerations. For example, precise species identification is critical for enforcing regulations related to the trade of endangered or protected species and for promoting sustainable forestry practices.

Traditionally, the classification of wood has been performed using macroscopic and microscopic techniques by trained experts~\cite{Desch1984, IAWA1989}. Macroscopic analysis typically involves observing surface features such as grain pattern, color, and texture, whereas microscopic examination requires the study of anatomical details like vessel elements, ray structures, and parenchyma distribution under magnification. These analyses are often guided by standardized keys and atlases and demand significant experience and expertise to execute effectively.

 While these conventional methods are time-tested and continue to be indispensable in many contexts, they come with several notable limitations. They are inherently labor-intensive and time-consuming, often requiring careful sample preparation and prolonged examination times.  A key limitation is their reliance on the skill and objectivity of the examiner, which introduces subjectivity into the identification process \cite{Wheeler2011, Gasson2011,Liu2022}. Interpretations based on subtle anatomical features can vary between experts, leading to inconsistency and reduced reliability \cite{Gasson2011}.

This subjectivity is further exacerbated by natural intra-species anatomical variability. Factors such as tree age, growth conditions, and the position of the sample within the tree (e.g., heartwood versus sapwood) can cause significant overlap in diagnostic features across species \cite{Gasson2010, Gasson2011}. As a result, even experienced wood anatomists may disagree when attempting to distinguish between closely related or visually similar species \cite{Gasson2010, UNODC2016}.

Forensic and regulatory guidelines also recognize the challenges posed by inter-observer discrepancies and sample degradation, which complicate consistent identification \cite{Gasson2011, UNODC2016}. These issues highlight the pressing need for more objective, reproducible, and accessible approaches to wood identification \cite{RosaDaSilva2022}. 
Moreover, these methods may be impractical in remote or resource-limited settings, where access to expert knowledge, reference materials, or specialized equipment is often lacking \cite{Richter2019}.

In response to these challenges, there has been growing interest in applying data-driven approaches—particularly those based on machine learning and computer vision—to automate and enhance wood classification tasks. In recent years, the rise of deep learning has transformed the landscape of artificial intelligence and has demonstrated remarkable success in a variety of complex recognition problems~\cite{LeCun2015,song2024,song2025,song_bioacoustic_generation2025,zhang2023dive}. Convolutional neural networks (CNNs), a class of deep learning models designed specifically for image data, have proven especially powerful due to their ability to automatically learn and hierarchically extract features from raw images~\cite{Krizhevsky2012}.

The field of wood identification has begun to benefit from these technological advances. Several studies have explored the application of deep learning to this domain, showing encouraging results in both laboratory and real-world settings~\cite{Ravindran2018,Oktaria2019,HerreraPoyatosIJCNN2024,Zielinski2025}. These approaches typically involve training a CNN on a labeled dataset of wood images, allowing the model to learn the distinguishing visual patterns associated with each species. Compared to traditional methods, deep learning offers the potential for greater speed, consistency, and scalability, as well as reduced reliance on expert knowledge during the classification process.

The motivation for the current study lies in harnessing these advantages for wood classification tasks in the context of Vietnamese forestry. Vietnam possesses rich biodiversity and is home to many commercially and ecologically significant wood species. However, tools for accurate and efficient species identification remain limited, especially in field or industrial settings where fast decision-making is crucial. By leveraging deep learning, we aim to develop a robust and scalable image-based classification system that can aid in species recognition and promote sustainable wood usage.

In this paper, we investigate the feasibility and effectiveness of applying deep learning models to the classification of wood images collected from ten tree species in Vietnam. We compare the performance of six widely used CNN architectures and assess their suitability for practical deployment. Our contributions include the creation of a curated image dataset, an empirical evaluation of multiple deep learning models, and insights into their performance on a real-world wood classification task.

Our evaluation demonstrates that lightweight convolutional neural network architectures can achieve high classification accuracy—exceeding 99\%—while maintaining low computational cost, making them suitable for deployment in resource-constrained environments such as mobile and field applications. Among the tested models, ShuffleNetV2 stood out for its optimal balance of performance and efficiency. These findings highlight the practical potential of deep learning to advance automated wood species identification and support sustainable forestry management in Vietnam.

The remainder of this paper is structured as follows. Section~\ref{datasection} describes the dataset and preprocessing steps. Section~\ref{methods} outlines the deep learning architectures used, as well as the computing environment and training configuration. Section~\ref{results} presents the classification results and analysis. Finally, Section~\ref{conclusion} summarizes our findings and discusses future research directions.

\section{Dataset and Preprocessing} \label{datasection}
In this section, we  describe the dataset construction and preparation workflow. We first detail the data collection process involving ten tree species from Vietnam. Then, we give a description of preprocessing and data augmentation techniques used to prepare the dataset for training.

\subsection{Data Collection}
Wood samples were collected from ten tree species commonly planted in the northern mountainous region of Vietnam: \textit{Michelia tonkinensis} (D), \textit{Nauclea orientalis} (G), \textit{Acacia hybrid} (KL), \textit{Acacia mangium} (KTT), \textit{Chukrasia tabularis} (LH), \textit{Erythrophleum fordii} (LX), \textit{Manglietia conifera} (M), \textit{Cinnamomum cassia} (Q),  and \textit{Melia azedarach} (XT)  from Thai Nguyen province, and \textit{Tectona grandis} (T) from Son La province, Vietnam. These plantation forests are owned and managed by smallholders.  The average annual rainfall in Thai Nguyen and Son La is 1{,}484~mm and 1{,}802~mm, respectively, while the average annual temperatures are 23.64\textdegree C and 24.06\textdegree C.

For each species, five individual trees were selected based on straightness, normal branching, and showing of no signs of any diseases or pest symptoms.  The north and south sides of the sample trees were marked before felling.

Detailed information on the sampling locations, tree diameter at breast height (DBH), total height, and estimated age is provided in Table~\ref{tab:species_info}.

\begin{figure}[H]
  \centering
  \begin{subfigure}[b]{0.3\textwidth}
    \includegraphics[width=\textwidth]{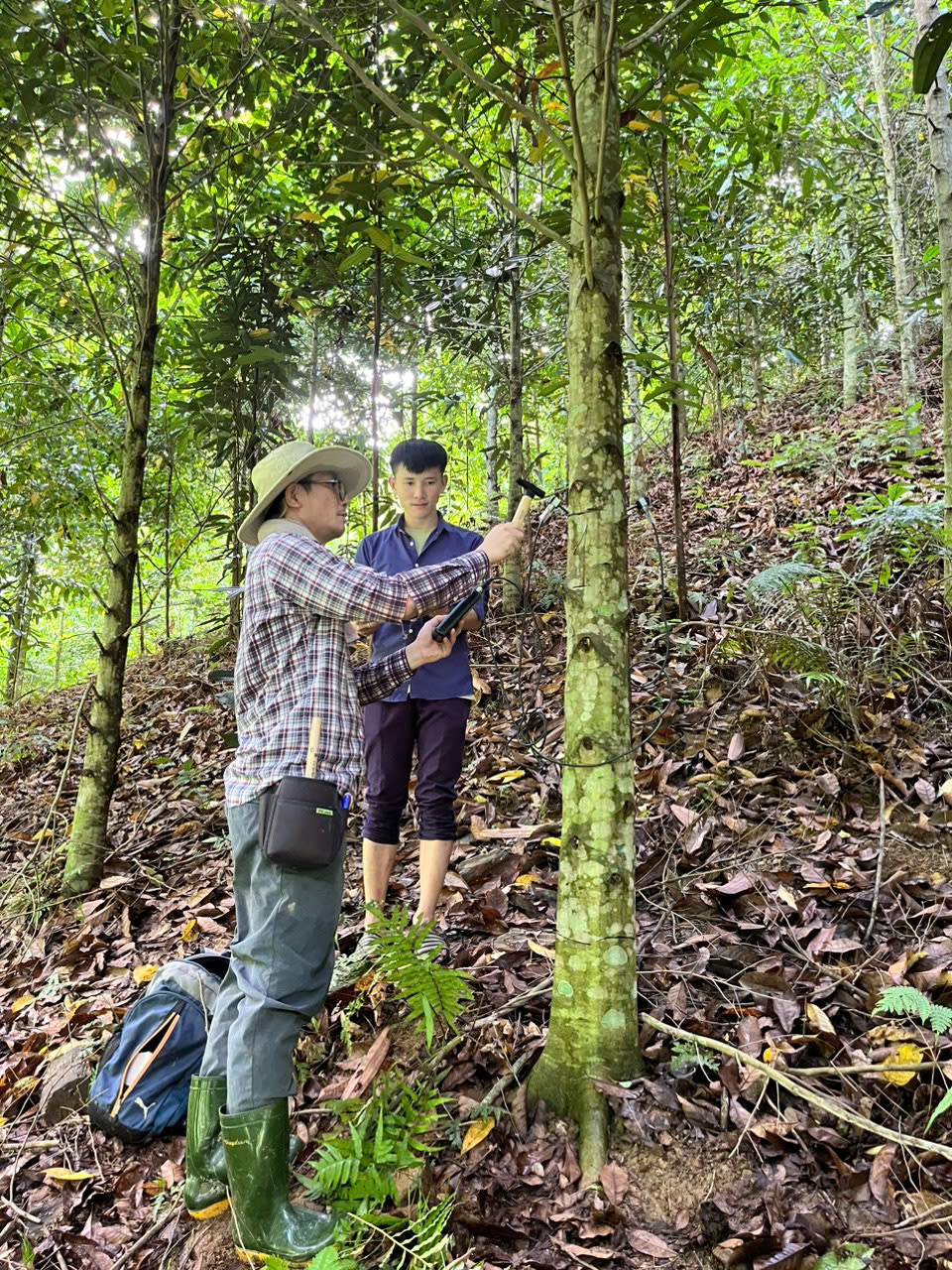}
  \end{subfigure}
  \begin{subfigure}[b]{0.3\textwidth}
    \includegraphics[width=\textwidth]{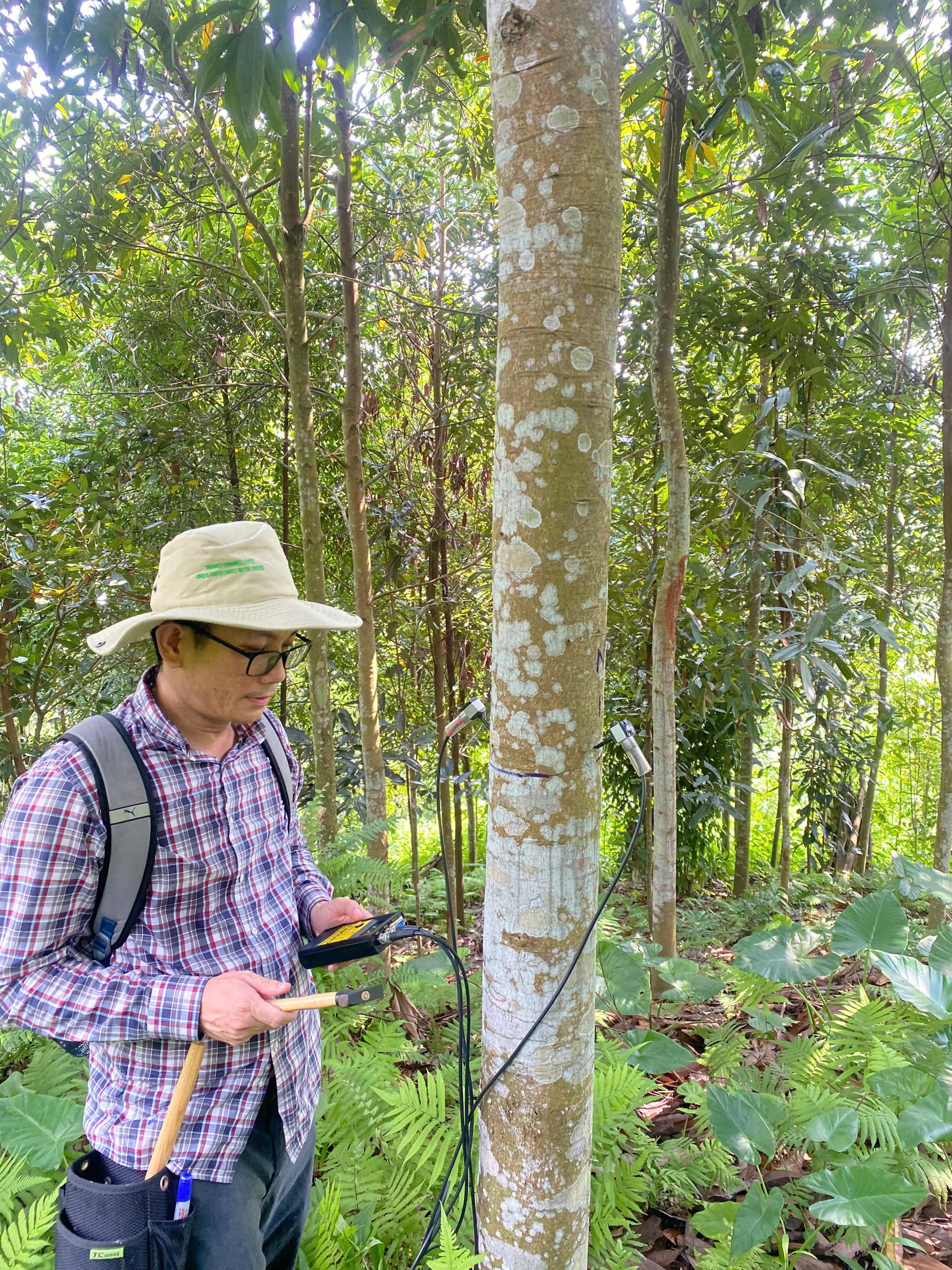}
  \end{subfigure}
  \begin{subfigure}[b]{0.3\textwidth}
    \includegraphics[width=\textwidth]{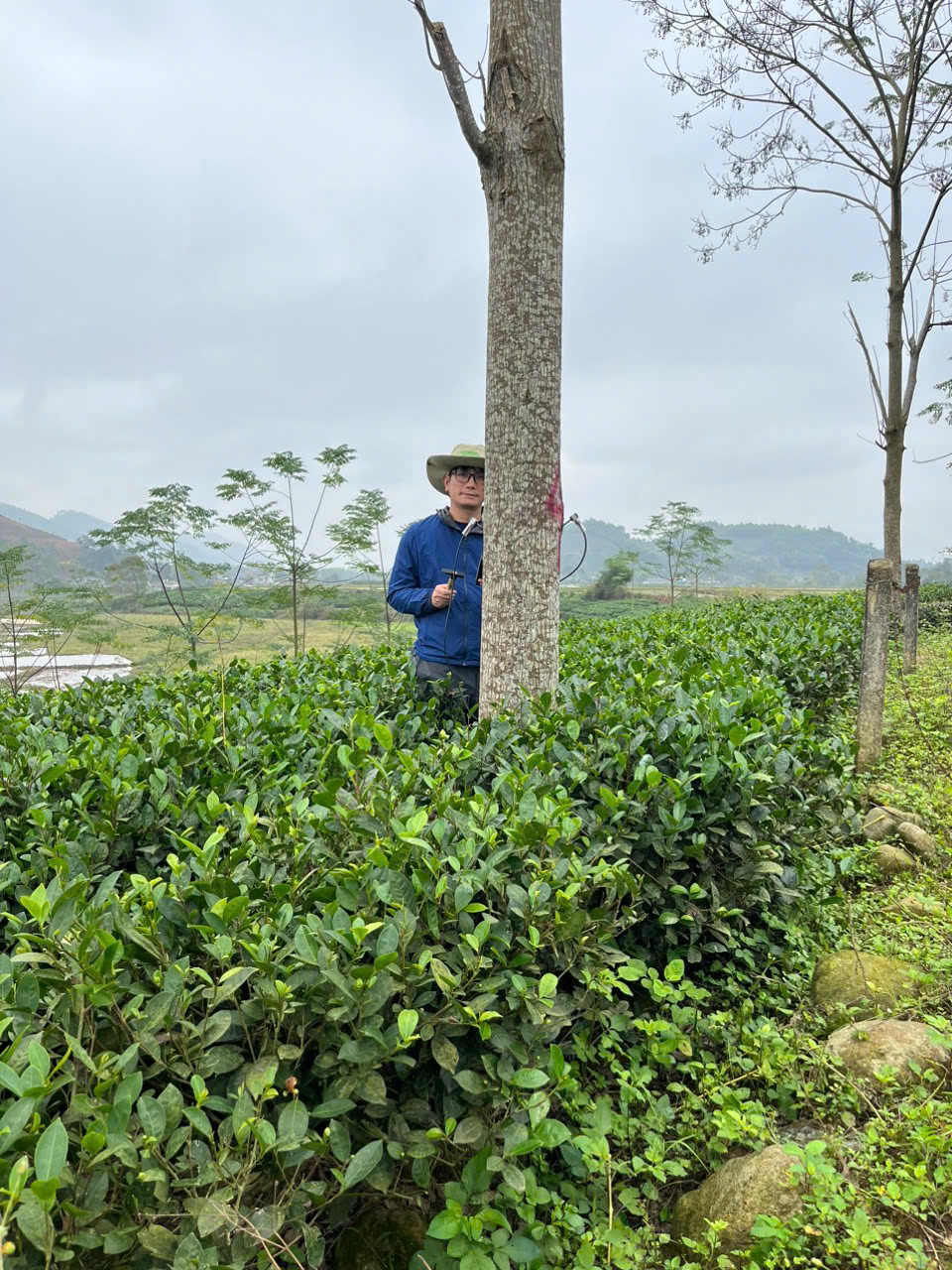}
  \end{subfigure}

  \begin{subfigure}[b]{0.3\textwidth}
    \includegraphics[width=\textwidth]{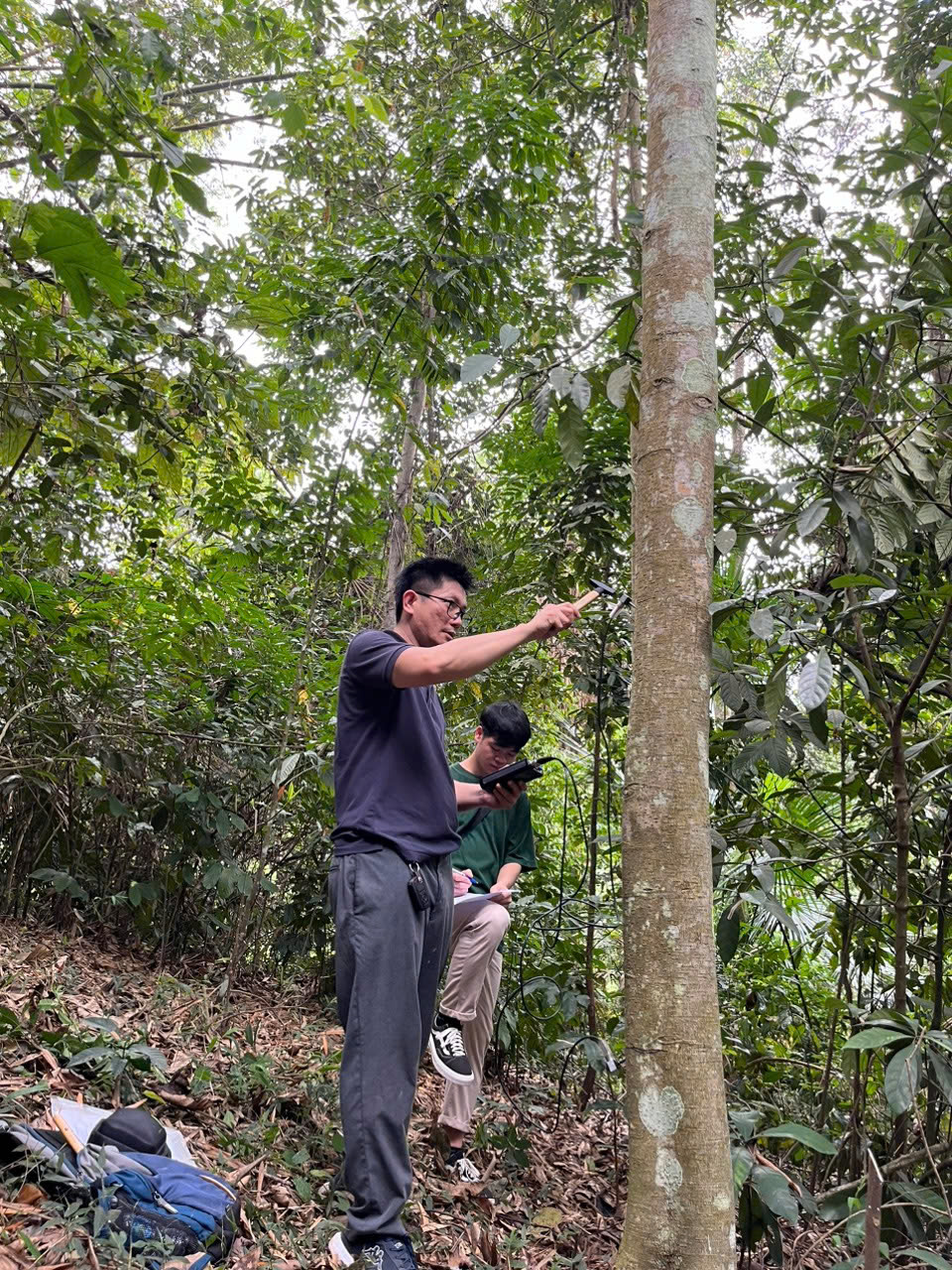}
  \end{subfigure}
  \begin{subfigure}[b]{0.3\textwidth}
    \includegraphics[width=\textwidth]{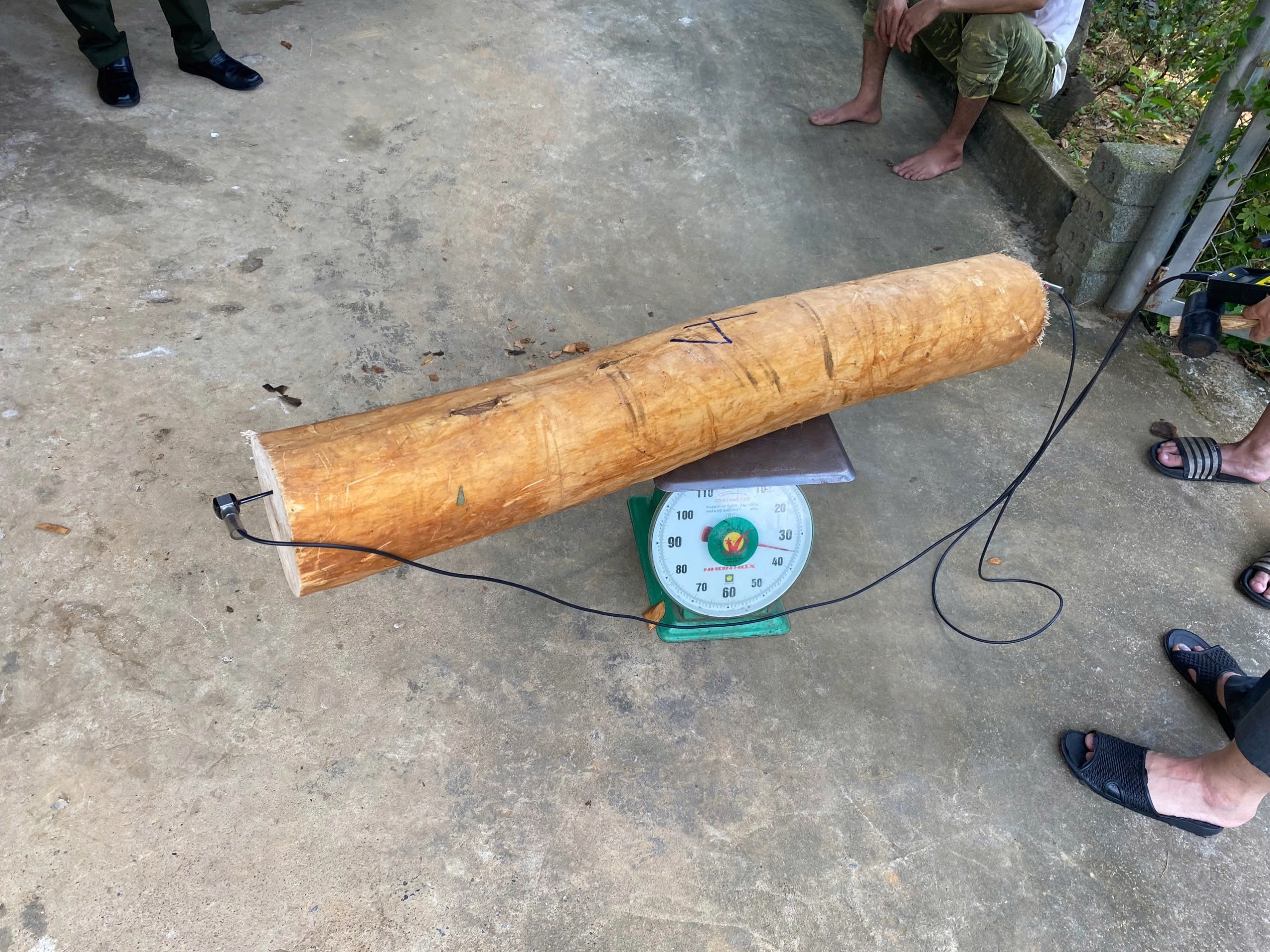}
  \end{subfigure}

  \caption{Photos of sampled trees collected during fieldwork.}
  \label{fig:tree_photos}
\end{figure}

\begin{table}[H]
\centering
\caption{Fundamental information of ten wood species samples}
\label{tab:species_info}
\begin{tabular}{p{4cm}llccc}
\toprule
\textbf{Location} & \textbf{Tree} & \textbf{n} & \textbf{DBH (cm)} & \textbf{Tree Height (m)} & \textbf{Age (year)} \\
\midrule
My Yen commune,\\Dai Tu district, Thai Nguyen & XT & 5 & 26.55 ± 1.66 & 14.02 ± 0.71 & 8 \\
My Yen commune,\\Dai Tu district, Thai Nguyen & KTT & 5 & 22.24 ± 1.54 & 10.72 ± 1.01 & 7 \\
My Yen commune,\\Dai Tu district, Thai Nguyen & LX & 5 & 21.02 ± 1.54 & 7.66 ± 0.48 & 17 \\
\midrule
Yen Ninh commune,\\Phu Luong district, Thai Nguyen & KL & 5 & 20.18 ± 0.30 & 13.56 ± 0.45 & 7 \\
Yen Ninh commune,\\Phu Luong district, Thai Nguyen & M & 5 & 20.31 ± 0.94 & 10.86 ± 0.58 & 14 \\
Yen Ninh commune,\\Phu Luong district, Thai Nguyen & LH & 5 & 22.26 ± 0.73 & 9.86 ± 0.47 & 27 \\
\midrule
Kim Phuong commune,\\Dinh Hoa district, Thai Nguyen & G & 5 & 21.78 ± 0.69 & 14.58 ± 0.64 & 15 \\
Kim Phuong commune,\\Dinh Hoa district, Thai Nguyen & Q & 5 & 15.57 ± 0.36 & 12.10 ± 0.43 & 13 \\
\midrule
Cay Thi commune,\\Dong Hy district, Thai Nguyen & D & 5 & 13.86 ± 0.72 & 9.42 ± 0.24 & 5 \\
\midrule
Chieng Hac commune,\\Yen Chau district, Son La & T & 5 & 22.21 ± 0.58 & 19.02 ± 0.44 & 23 \\
\bottomrule
\end{tabular}
\end{table}

From each tree,  we cut a 1.0 m log horizontally at a height of 0.3 m to 1 m above ground level. After air-drying in a room at ambient conditions (no humidity control) for approximately 2 months, a cross-sectional disc 100 mm thick was cut from each log at a height 1.3 m. From each disc, samples with dimensions of 80 (radial) $\times$ 15 (tangential) $\times$ 100 (longitudinal) mm$^3$ were carefully cut from each cardinal direction (north--south and east--west) to capture the cross-section of the wood.  As a result, all subsamples included both heartwood and sapwood regions of the tree.

Photographs of all subsamples were taken using an iPhone 13 (Apple Inc., USA) with 2x zoom, equipped with an APEXEL 200X high-resolution macro lens. The images were captured under natural morning light near a sunlit window, with backlighting to enhance visibility. The photographs covered various surface regions of each sample, highlighting differences in grain pattern, color, and texture to increase dataset diversity and realism.

Our dataset comprises a total of 29,292 wood images. Table~\ref{tab:tree_counts} details the distribution of images per species.

\begin{table}[H]
\centering
\begin{tabular}{lll lll}
\toprule
\textbf{Tree} & \textbf{Number} & \textbf{Tree} & \textbf{Number} & \textbf{Tree} & \textbf{Number} \\
\midrule
KTT & 4,137 & Q   & 3,440 & G   & 3,432 \\
M   & 3,424 & LH  & 3,365 & KL  & 3,319 \\
XT  & 3,271 & LX  & 3,080 & D   & 1,266 \\
T   & 558   &      &       &     &       \\
\bottomrule
\end{tabular}
\caption{Number of images per tree species in the dataset.}
\label{tab:tree_counts}
\end{table}

\subsection{Data Augmentation and Preprocessing}
To enhance model generalization and mitigate overfitting, data augmentation techniques were applied to the training images. Before augmentation, the dataset was split into training, validation, and test subsets in a ratio of approximately 60\%, 20\%, and 20\%, respectively. All images were resized to a fixed resolution of $224 \times 224$ pixels to ensure compatibility with standard convolutional neural network architectures.

Data augmentation operations applied to the training set included random horizontal flipping, random rotations within a range of $\pm 15^{\circ}$, brightness and contrast adjustments, and the addition of Gaussian noise. These transformations simulate real-world variability and help the models learn more robust features. Importantly, no augmentation was applied to the validation or test sets, preserving them for unbiased evaluation of model performance.

\begin{figure}[h]
\centering
\includegraphics[width=120mm]{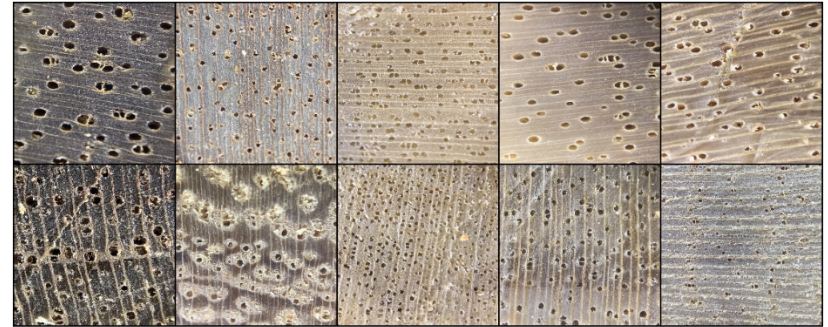}
\caption{Examples of training images after data augmentation.}
\label{fig:augmentation}
\end{figure}

Figure~\ref{fig:augmentation} illustrates examples of augmented images from the training set.

\section{Methods}  \label{methods}
 In this section, we  begin by introducing five deep learning models selected for our study, along with the rationale behind their selection. Next, we describe the training setup and hyperparameter configurations. Finally, we outline the evaluation metrics used to assess model performance.

\subsection{Model Selection}
 To evaluate the performance of deep learning models for wood species classification, we adopted a selection strategy based on three key considerations. First, we prioritized models that are suitable for real-world deployment, particularly those optimized for resource-constrained environments such as mobile or embedded systems. Second, to avoid redundancy, we selected only one representative model from families with similar architectural foundations. For instance, among deep residual networks, we included ResNet while omitting closely related architectures like DenseNet and VGG. Likewise, among transformer-based models, we chose MobileViT due to its efficiency and suitability for lightweight applications, rather than including other variants such as Swin Transformer. Third, we focused on well-established and widely adopted architectures, excluding some recently proposed models that, while achieving state-of-the-art results in niche domains, lack broad validation or deployment readiness.

Based on these criteria,  we selected five widely used and representative architectures: ResNet50~\cite{he2016deep}, EfficientNet~\cite{tan2019efficientnet}, MobileViT~\cite{mehta2021mobilevit}, MobileNetV3~\cite{howard2019searching}, and ShuffleNetV2~\cite{ma2018shufflenet}. These models encompass a diverse range of design philosophies, spanning from classical convolutional neural networks (CNNs) to recent hybrid and transformer-based architectures, as well as from heavyweight models with high representational capacity to lightweight models optimized for resource-constrained deployment.

Let us now give a   brief overview of each selected model.

\textbf{ResNet50:} ResNet (Residual Network) introduces residual learning to mitigate the vanishing gradient problem in very deep networks. The 50-layer ResNet50 variant provides robust feature extraction and has become a standard baseline for image classification tasks. However, its relatively large parameter count and computational cost may limit its suitability for real-time or mobile applications.

\textbf{EfficientNet:} EfficientNet employs a compound scaling strategy that jointly scales network depth, width, and input resolution, achieving a superior balance between accuracy and efficiency. While EfficientNet performs competitively on standard benchmarks, it still requires substantial computational resources, particularly for training with high-resolution images.

\textbf{MobileViT:} MobileViT combines convolutional layers with transformer blocks to leverage both local feature extraction and global context modeling. This hybrid architecture captures long-range dependencies more effectively than traditional CNNs while maintaining a relatively small parameter footprint. MobileViT is designed for lightweight vision tasks and offers a favorable trade-off between accuracy and efficiency.

\textbf{MobileNetV3:} The latest iteration in the MobileNet series, MobileNetV3 integrates advanced components such as inverted residuals, squeeze-and-excitation modules, and hard-swish activations. It is specifically optimized for mobile and embedded platforms, delivering strong accuracy with significantly reduced parameters and computational demand.

\textbf{ShuffleNetV2:} Tailored for high-speed and low-power environments, ShuffleNetV2 utilizes channel splitting and element-wise operations to minimize computation and memory access costs. It improves upon its predecessor by addressing memory bottlenecks and enhancing efficiency, making it well-suited for mobile deployment, though potentially at a modest cost to classification accuracy.

The selected models represent a broad spectrum of computational requirements and design principles, allowing for a comprehensive evaluation of their effectiveness and practicality for wood species classification tasks.

\subsection{Training Setup and Hyperparameters}
All models were trained from scratch on our custom wood image dataset without employing transfer learning or pretraining on external datasets such as ImageNet. This approach ensures that the learned features are fully tailored to the specific characteristics of our wood classification task and not influenced by unrelated visual domains.

Training was conducted using the PyTorch framework on three different systems:

\begin{itemize}
  \item \textbf{System 1:} A CERVO-GRASTA workstation featuring an Intel Xeon W-2255 CPU, 256GB DDR4-2933 ECC RAM, 1.92TB NVMe SSD, 8TB HDD, and an NVIDIA GeForce RTX 3090 GPU (24GB GDDR6X).
  \item \textbf{System 2:} A high-performance server powered by an Intel Core i9-10850K CPU (10 cores, up to 5.2GHz), 64GB RAM, and an NVIDIA GeForce RTX 3090 GPU.
  \item \textbf{System 3:} A workstation equipped with an NVIDIA GeForce RTX 3060 GPU (12GB VRAM), running Ubuntu 20.04 LTS with CUDA 11.5.
\end{itemize}

The Adam optimizer was used for all training sessions with an initial learning rate of $5 \times 10^{-4}$. We used a batch size of 64, and categorical cross-entropy was adopted as the loss function for this multi-class classification problem. Each model was trained for 50 epochs. Early stopping was not applied to allow for consistent and comparable learning behavior across all models. Model checkpoints were saved at regular intervals for further analysis.

All input images were resized to $224 \times 224$ pixels to match the input requirements of standard CNN architectures. Data augmentation techniques, as described in Section~\ref{datasection}, were applied exclusively to the training set to improve generalization and reduce overfitting. No augmentation was applied to the validation or test sets to ensure a fair and unbiased evaluation.

The next section presents the evaluation metrics used to assess model performance.

\subsection{Evaluation Metrics}

We evaluate model performance using a combination of classification and efficiency metrics, including accuracy, loss, recall, F1-score, model size, floating-point operations per second (FLOPs), and inference latency. These metrics allow for evaluating not only classification effectiveness but also computational efficiency—an important factor for field-deployable systems.

For completeness, we briefly restate the recall and F1-score metrics used in our analysis.

Recall, also known as the true positive rate, measures the proportion of actual positive samples that were correctly identified by the model. It is defined as:
\[
\text{Recall} = \frac{\text{True Positives}}{\text{True Positives} + \text{False Negatives}}.
\]
High recall indicates that the model successfully detects most of the relevant instances for each wood species.

Meanwhile, the F1-score, a widely used metric in classification tasks, is the harmonic mean of precision and recall. It provides a single value that balances the trade-off between false positives and false negatives, and is especially informative when the dataset is imbalanced. The F1-score is defined as:
\[
\text{F1-score} = 2 \cdot \frac{\text{Precision} \cdot \text{Recall}}{\text{Precision} + \text{Recall}}.
\]
A high F1-score indicates that a model achieves both strong precision and recall, thereby offering a more comprehensive measure of classification quality than accuracy alone.

\section{Results}  \label{results}

This section presents a comprehensive evaluation of the selected models, focusing on both classification performance and computational efficiency. We report standard classification metrics—including accuracy, recall, and F1-score—alongside model size, FLOPs, and inference latency to assess deployment feasibility.

We begin with the training and validation performance of all models in Subsection~\ref{ValidationPerformance}, including losses and accuracies. Subsection~\ref{ClassificationAccuracy} presents classification results on the test set, focusing on accuracy and recall. Since these results are based on single training runs, we further assess model robustness through statistical evaluation in Subsection~\ref{StatisticalEvaluation}, where we analyze the mean and standard deviation of accuracy, recall, and F1-score over multiple runs.

In Subsection~\ref{ComputationalEfficiency}, we evaluate the practical deployment feasibility of each model by comparing the number of parameters, FLOPs, and inference latency. Based on the findings from the four subsections above, we provide an overall comparison in Subsection~\ref{OverallComparison} to identify the most suitable model. Finally, we analyze the confusion matrix of the best-performing model to examine its per-class classification behavior.

\subsection{Training and Validation Performance} \label{ValidationPerformance}
Table~\ref{table_training} summarizes the training and validation performance of all models. Each model achieved high training accuracy, indicating effective learning during training. Among them, ShuffleNetV2 recorded the best results across three metrics: lowest training loss (0.0074), highest training accuracy (99.80\%), and lowest validation loss (0.0052). Its validation accuracy (99.83\%) was the second-highest, closely following EfficientNet’s top score of 99.91\%.

MobileNetV3 showed slightly higher losses and lower accuracies during both training and validation compared to the other models, though it still achieved satisfactory performance overall.

\begin{table}[H]
\centering
\small
\setlength{\tabcolsep}{3pt}
\begin{tabular}{lcccc}
\toprule
\textbf{Model} & \textbf{Training Loss} & \textbf{Training Accuracy} & \textbf{Validation Loss} & \textbf{Validation Accuracy} \\
\midrule
ResNet50 & 0.0145 & 99.50\% & 0.0168 & 99.40\% \\
EfficientNet & 0.0135 & 99.61\% & 0.0028 & \textbf{99.91\%} \\
MobileViT & 0.0117 & 99.62\% & 0.0252 & 99.32\% \\
MobileNetV3 & 0.0228 & 99.27\% & 0.0239 & 99.21\% \\
ShuffleNetV2 & \textbf{0.0074} & \textbf{99.80\%} & \textbf{0.0052} & 99.83\% \\
\bottomrule
\end{tabular}
\caption{Training and validation loss/accuracy of each model.}
\label{table_training}
\end{table}

\subsection{Test Set Classification Accuracy}  \label{ClassificationAccuracy}
The performance of each model on the held-out test set is reported in Table~\ref{table_acc}. All models achieved strong performance, with test accuracy and recall above 98\%.



Among models, EfficientNet achieved the highest accuracy (99.84\%) and recall (99.78\%). ShuffleNetV2 performed comparably, attaining 99.79\% recall and 99.78\% accuracy—despite its lower complexity. MobileViT, a hybrid CNN-transformer architecture, also performed well with the highest recall (99.81\%).

MobileNetV3 yielded the lowest test performance, though still strong, with accuracy of 98.55\% and recall of 98.41\%. These differences, while small, could be influenced by intra-class visual similarity or class imbalance. To confirm robustness, we include a statistical evaluation in Subsection \ref{StatisticalEvaluation}.

\begin{table}[H]
\centering
\begin{tabular}{lcc}
\toprule
\textbf{Model} & \textbf{Recall} & \textbf{Accuracy} \\
\midrule
ResNet50 & 99.20\% & 99.31\% \\
EfficientNet & 99.78\% & \textbf{99.84\%} \\
MobileViT & \textbf{99.81\%} & 99.77\% \\
MobileNetV3 & 98.41\% & 98.55\% \\
ShuffleNetV2 & 99.79\% & 99.78\% \\
\bottomrule
\end{tabular}
\caption{Classification accuracy and recall on the test set.}
\label{table_acc}
\end{table}



\subsection{Statistical Evaluation}  \label{StatisticalEvaluation}

To evaluate the robustness and consistency of each model, we conducted 20 independent training runs per architecture using different random seeds. The full set of experiments required approximately 12 days to complete on our workstations. All experimental settings, including data splits and augmentation strategies, were held constant across runs. Table~\ref{tab:statistical_results} presents the mean and standard deviation of three key performance metrics: accuracy, recall, and F1-score.

Among all models, ShuffleNetV2 demonstrated the most stable and highest overall performance, achieving an average accuracy of \textbf{99.29\% $\pm$ 0.15}, recall of \textbf{99.38\% $\pm$ 0.12}, and F1-score of \textbf{99.35\% $\pm$ 0.13}. EfficientNet and MobileViT also showed consistent and high performance across runs. In contrast, MobileNetV3 exhibited greater variability and slightly lower average scores, reflecting its trade-off between computational efficiency and classification accuracy.

\begin{table}[H]
\centering
\begin{tabular}{lccc}
\toprule
\textbf{Model} & \textbf{Accuracy (\%)} & \textbf{Recall (\%)} & \textbf{F1-score (\%)} \\
\midrule
ResNet50 & 96.78 $\pm$ 1.65 & 97.08 $\pm$ 1.33 & 96.72 $\pm$ 1.70 \\
EfficientNet & 99.21 $\pm$ 0.24 & 99.14 $\pm$ 0.43 & 99.12 $\pm$ 0.42 \\
MobileViT & 99.03 $\pm$ 0.25 & 98.90 $\pm$ 0.35 & 98.86 $\pm$ 0.33 \\
MobileNetV3 & 94.34 $\pm$ 1.94 & 94.44 $\pm$ 1.82 & 94.33 $\pm$ 1.93 \\
ShuffleNetV2 & \textbf{99.29 $\pm$ 0.15} & \textbf{99.38 $\pm$ 0.12} & \textbf{99.35 $\pm$ 0.13} \\
\bottomrule
\end{tabular}
\caption{Statistical evaluation across 20 independent runs (mean $\pm$ standard deviation).}
\label{tab:statistical_results}
\end{table}

\subsection{Computational Efficiency}  \label{ComputationalEfficiency}

To evaluate deployment feasibility, Table~\ref{table_com} summarizes the number of parameters,  FLOPs, and inference latency. ResNet50 had the highest parameter count (23.53M) and latency (33.54ms), whereas MobileNetV3 was the most lightweight, with only 1.53M parameters and the fastest inference (7.84ms).

Despite its low complexity, ShuffleNetV2 maintained top-tier accuracy with only 1.26M parameters and 13.06ms latency, making it ideal for mobile or embedded systems.

\begin{table}[H]
\centering
\begin{tabular}{lccc}
\toprule
\textbf{Model} & \textbf{Parameters} & \textbf{FLOPs} & \textbf{Latency} \\
\midrule
ResNet50 & 23.53M & 4.09G & 33.54ms \\
EfficientNet & 4.02M & 0.38G & 18.66ms \\
MobileViT & 0.95M & 0.15G & 15.09ms \\
MobileNetV3 & 1.53M & \textbf{0.06G} & \textbf{7.84ms} \\
ShuffleNetV2 & 1.26M & 0.14G & 13.06ms \\
\bottomrule
\end{tabular}
\caption{Computational efficiency metrics for all models.}
\label{table_com}
\end{table}



\subsection{Overall Comparison and Confusion Matrix Analysis}  \label{OverallComparison}
Balancing accuracy, consistency, and computational efficiency, ShuffleNetV2 emerges as the most promising model for practical deployment. It achieves state-of-the-art accuracy and robustness with minimal resource requirements, making it ideal for ecological fieldwork or mobile applications.

\begin{figure}[H]
\centering
\includegraphics[width=100mm]{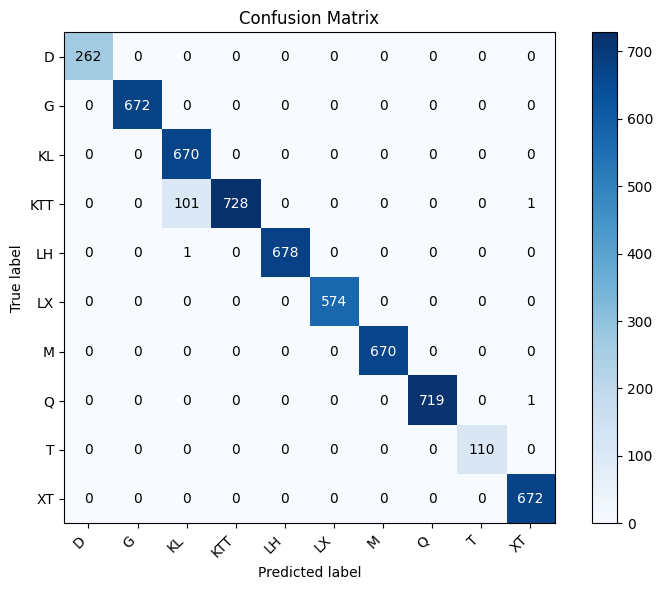}
\caption{Confusion matrix for ShuffleNetV2 predictions.}
\label{fig:cm}
\end{figure}

To analyze per-class performance, we calculate the confusion matrix  of this best-performing model.  Figure~\ref{fig:cm} shows the confusion matrix for ShuffleNetV2. 
The diagonal elements indicate correctly classified instances for each species, while off-diagonal elements correspond to misclassifications.  As shown, most species were classified with 100\% accuracy, indicating minimal class confusion. However, minor  misclassification occurred between \textit{Acacia hybrid} (KL) and \textit{Acacia mangium} (KTT), which is likely attributable to their high visual similarity, as both species belong to the same \textit{Acacia}  genus and exhibit overlapping anatomical features.

These results suggest that although the model is highly accurate overall, improving inter-species discrimination—particularly for visually similar classes—may require additional features or higher-resolution imaging.

\section{Conclusion}  \label{conclusion}
In this study, we conducted a comprehensive evaluation of several deep learning models for the task of wood species classification using a curated dataset of ten common species found in Vietnam. The models assessed include convolutional neural networks (ResNet50, EfficientNet, ShuffleNetV2, MobileNetV3) and a hybrid architecture (MobileViT).

Among all models, ShuffleNetV2 achieved the best overall performance, combining high classification accuracy (99.78\%) and recall (99.79\%) with excellent computational efficiency. Its small parameter size (1.26M), low computational cost (0.14 G FLOPs), and fast inference time (13.06 ms) make it particularly well-suited for deployment in resource-limited environments, such as mobile or field-based ecological monitoring systems.

MobileViT also showed competitive results, striking a strong balance between model size and classification performance. However, its slightly higher inference latency may be a limitation in time-sensitive applications. MobileNetV3, while not achieving the top classification metrics, remains a viable candidate for ultra-low-latency scenarios, thanks to its minimal computational footprint.

Overall, our findings indicate that lightweight CNN-based models continue to offer the best trade-off between accuracy and efficiency for ecological image classification tasks, especially when deployed in environments with limited hardware resources. 

Future work will focus on enhancing performance through fine-tuning techniques, improved augmentation strategies, and ensemble modeling. In addition, expanding the dataset to include more wood species and incorporating multispectral or microscopic imaging could further improve model robustness and generalizability.

\section*{Acknowledgement}
The last author (T.V.T.) was supported by the Tokuo Fujii Research Fund -- Support for Article Processing Charge of International Academic Papers. The second author (V.D.D) was supported by the Thai Nguyen Department of Science and Technology (Grant No. DT/KTCN/26/2023) and the Vietnam Ministry of Education and Training (Grant No. B2024-TNA-13).

\end{document}